\documentclass{article}

\usepackage{microtype}
\usepackage{graphicx}
\usepackage{subfigure}
\usepackage{booktabs} 

\usepackage{hyperref}

\usepackage[accepted]{icml2025}

\usepackage{amsmath}
\usepackage{amssymb}
\usepackage{mathtools}
\usepackage{amsthm}

\usepackage[capitalize,noabbrev]{cleveref}

\theoremstyle{plain}

\theoremstyle{definition}

\theoremstyle{remark}

\usepackage{tcolorbox}
\usepackage{xspace}
\newcommand{\method}{\textit{ReasonFlux}\xspace}
\usepackage[textsize=tiny]{todonotes}

\usepackage{tcolorbox}
\icmltitlerunning{ReasonFlux: Hierarchical LLM Reasoning via Scaling Automated Thought Templates}

\begin{document}

\onecolumn
\icmltitle{ReasonFlux: Hierarchical LLM Reasoning via Scaling Thought Templates}



\icmlsetsymbol{equal}{*}

\begin{icmlauthorlist}
\icmlauthor{Ling Yang$^{\ 1}$}{equal}
\icmlauthor{Zhaochen Yu$^{\ 2}$}{equal}
\icmlauthor{Bin Cui$^{\ 2}$}{}
\icmlauthor{Mengdi Wang$^{\ 1}$}{}
\\
$^{1}$Princeton University\quad$^{2}$Peking University
\\
Code: \href{https://github.com/Gen-Verse/ReasonFlux}{https://github.com/Gen-Verse/ReasonFlux}
\end{icmlauthorlist}


\icmlcorrespondingauthor{Ling Yang}{yangling0818@163.com}
\icmlcorrespondingauthor{Mengdi Wang}{mengdiw@princeton.edu}

\icmlkeywords{Machine Learning, ICML}

\vskip 0.3in



\printAffiliationsAndNotice{\icmlEqualContribution} 

\begin{abstract}
We present that hierarchical LLM reasoning via scaling \textit{thought templates} can effectively optimize the reasoning search space and outperform the mathematical reasoning capabilities of powerful LLMs like OpenAI o1-preview and DeepSeek V3. We train our \method-32B model with only 8 GPUs and introduces three innovations: (\textbf{i}) a structured and generic thought template library, containing \textbf{around 500 high-level thought templates} capable of generalizing to similar or relevant reasoning problems; (\textbf{ii}) performing hierarchical reinforcement learning on a sequence of thought templates instead of original long CoT data, optimizing a base LLM to plan out an optimal template trajectory for gradually handling complex problems; (\textbf{iii}) a brand new inference scaling system that enables \textit{hierarchical LLM reasoning} by adaptively scaling thought templates at inference time. With a template trajectory containing more explainable reasoning structures than DeepSeek-R1 and o3-mini, our \method-32B significantly advances math reasoning capabilities to state-of-the-art levels.
Notably, on the MATH benchmark, it achieves an accuracy of \textbf{91.2\%} and surpasses o1-preview by 6.7\%. 
On the USA Math Olympiad (AIME) benchmark, \method-32B solves an average of \textbf{56.7\%} of problems, surpassing o1-preview and DeepSeek-V3 by 27\% and 45\%, respectively. 
\end{abstract}

\begin{table}[h]
\centering

\label{tab:summary}
\begin{tabular}{lccccccc}
\toprule
Task  & \textbf{ReasonFlux}& \textbf{DeepSeek} & \textbf{OpenAI} & \textbf{OpenAI} & \textbf{QWQ}&\textbf{GPT} \\
 &\textbf{32B}  &  \textbf{V3} & \textbf{o1-preview} & \textbf{o1-mini} &\textbf{32B-preview} &\textbf{4o}\\
\midrule
MATH & \textbf{91.2} &90.2& 85.5 & 90.0& 90.6&76.6 \\
AIME 2024 & \textbf{56.7} &39.2& 44.6 & 56.7 &50.0&9.3 \\
Olympiad Bench & \textbf{63.3} &55.4& - & 65.3  &61.2&43.3 \\
GaokaoEn 2023 & \textbf{83.6} &-& 71.4 & 78.4 &65.3&67.5 \\
AMC2023 & \textbf{85.0} &80.0& 90.0 & 95.0 & -&47.5\\
\bottomrule
\end{tabular}
\caption{Performance Comparison on Various Math Reasoning Benchmarks (Pass@1 Accuracy)}
\end{table}

\section{Introduction}

Large Language Models (LLMs) have recently achieved remarkable progress, demonstrating exceptional capabilities in tackling complex reasoning tasks and even surpassing human experts in specific domains. For example, models such as OpenAI's O1  \citep{jaech2024openai}, Google's Gemini-2.0 \citep{gemini}, DeepSeek-V3 \citep{liu2024deepseek}, and Qwen-QwQ \citep{qwq-32b-preview} are at the forefront of this progress, characterized by their ability to emulate human reasoning through a slower, more deliberate thought process. These models leverage increased inference time to enhance reasoning accuracy. While they have unlocked substantial performance gains, more complex tasks such as mathematical problem solving in AIME, OlympiadBench \citep{he2024olympiadbench} and code in LiveCodeBench \citep{jain2024livecodebench}, which demand a more fine-grained search through a vast solution space and more delicate thought for each intricate reasoning step, thus still pose significant challenges.

\begin{figure*}[tp]
    \centering
    
    \includegraphics[width=1\textwidth]{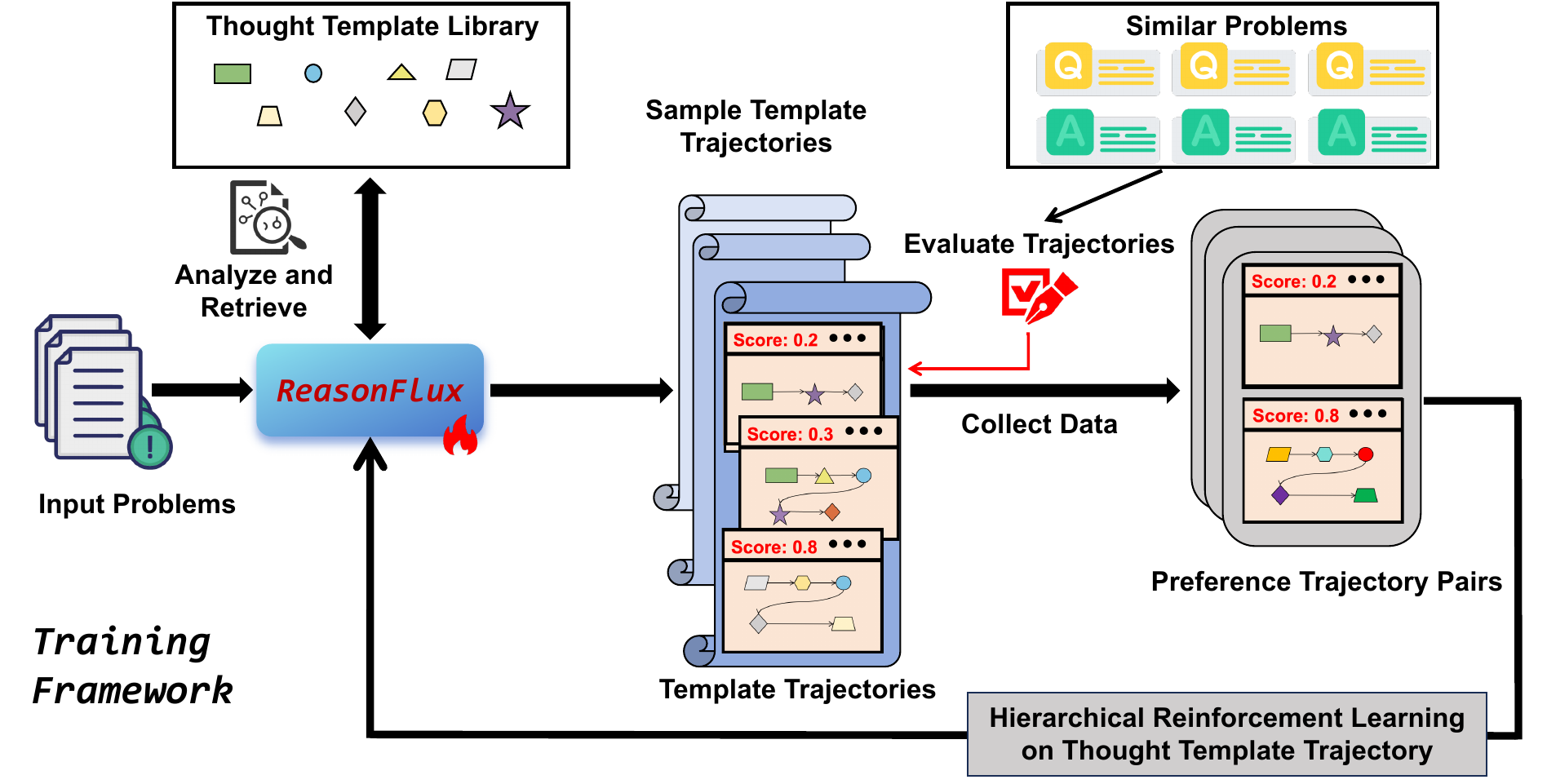}
    \caption{\textbf{Training framework for our \method.} We train with hierarchical reinforcement learning to enable the model to plan out an optimal and generalizable thought template trajectory for an input problem. Our new inference-scaling framework is in \cref{pic-inference}.}
    \label{pic-training}
\end{figure*}

Subsequent research has focused on enhancing LLMs' reasoning capabilities on complex problems through inference-time strategies. These strategies can be divided into two categories: deliberate search and reward-model-guided methods. Deliberate search methods, like Tree of Thoughts (ToT) \citep{ToT} and Graph of Thoughts (GoT) \citep{GoT}, allow LLMs to explore multiple reasoning paths and self-evaluate choices to find the optimal trajectory. Reward-model-guided methods leverage reward models to assess reasoning step quality. Best-of-N approaches, which leverage an Outcome Reward Model (ORM) to find the optimal reasoning paths in multiple candidates, while Process Reward Models (PRMs) \citep{lightman2023let,luo2024improve,wang2024math} guide the model towards promising paths by rewarding high-probability intermediate steps.
Building on this, Monte Carlo Tree Search (MCTS) \citep{zhang2024rest,qi2024mutual} employs a fine-grained search, decomposing tasks into simpler steps and using PRMs to guide action selection within a tree-based search space. However, these methods often incur high computational costs, especially with numerous reasoning steps or vast search spaces, primarily due to the inherent randomness of sampling, which hinders the efficient identification of the optimal reasoning trajectory.  Furthermore, they rely on manually designed search strategies and instance/step-level reward, limiting their generalization ability to diverse and complex reasoning tasks. Essentially, they struggle to effectively balance the exploration-exploitation trade-off  during inference scaling. This highlights the need for a more efficient and generalizable inference scaling approach that enhances reasoning without extensive manual effort, while providing a more principled search strategy.

To achieve more efficient and precise search of reasoning paths, a feasible approach is to utilize Retrieval-Augmented Generation (RAG). Recent Buffer of Thought (BoT) \citep{BoT} constructs a meta-buffer to store informative, high-level thoughts distilled from various problem-solving processes, adaptively retrieving and instantiating relevant thought templates for each specific task. 
SuperCorrect \citep{yang2024supercorrect} further utilizes both high-level and detailed thought templates to enhance reasoning ability of small LLMs. 
Despite significant improvements, such template-based reasoning methods may still face challenges when applied to complex reasoning tasks. Because complex problems often require the integration of multiple templates or diverse pieces of retrieved information, which current methods struggle to address effectively.

\begin{figure*}[tp]
    \centering
    \includegraphics[width=\textwidth]{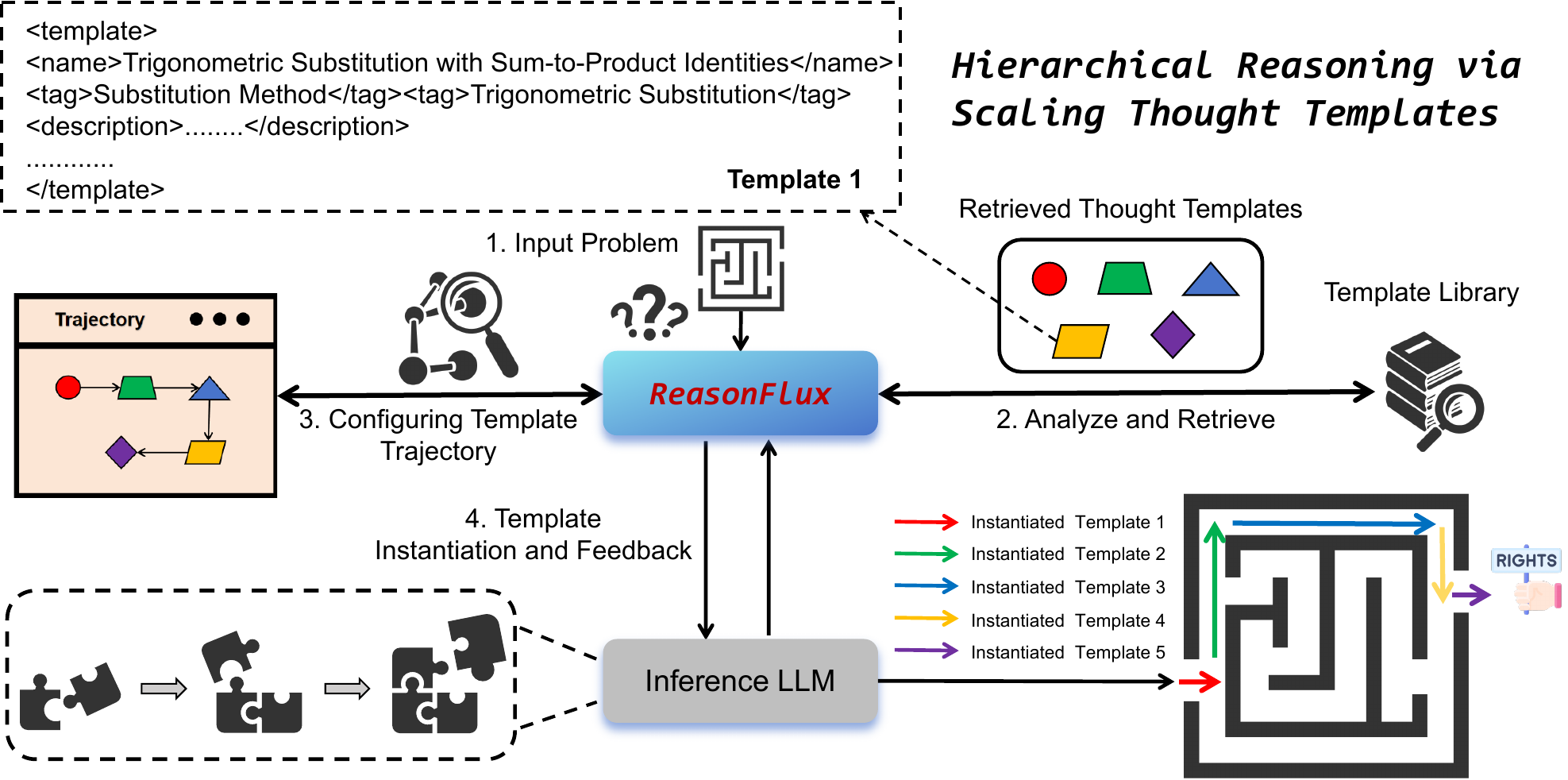}
    \caption{\textbf{New inference scaling system based on hierarchical reasoning.} We retrieve a series of high-level thought templates for complex problems, and gradually conduct instantiated reasoning for a sequence of sub-problems. }
    \vspace{-0.2in}
    \label{pic-inference}
\end{figure*}

To this end, we introduce \method, a novel hierarchical LLM reasoning framework that configures optimal \textit{thought template trajectories} by automatically retrieving relevant high-level thought templates at inference time, to achieve superior performance on complex reasoning tasks and even \textbf{outperform OpenAI o1-preview and o1-mini models}. To be more specific, we first construct a structured template library, which contains \textbf{500 useful compacted thought templates} for efficient retrieval and adaptation.  
Instead of optimizing a long CoT trajectory, we perform hierarchical reinforcement learning on a sequence of high-level thought templates, optimizing a base llm to learn an optimal \textbf{thought template trajectory} from multiple ones and guiding an inference LLM to solve a series of simpler sub-problems. 
Finally, we develop a new inference scaling system through adaptively \textbf{scaling thought templates}. This hierarchical reasoning paradigm enables \method to simplify the search of reasoning paths and enhance the reasoning ability for complex problems by dynamically selecting a most appropriate high-level template for each sub-problem.
Our automated template scaling allows \method to effectively achieve a better exploration-exploitation trade-off, leading to a more robust and efficient problem-solving process. 
Through these innovations, \method offers a more efficient, generalizable, and scalable solution for enhancing the complex reasoning capabilities of LLMs. Finally, we summarize our contributions as follows:

\begin{enumerate}
    \item We introduce \method (in \cref{pic-training}), a hierarchical LLM reasoning framework that significantly enhances complex reasoning capabilities, outperforming SOTA models like o1-preview and DeepSeek-V3 on challenging MATH and AIME benchmarks (in \cref{tab:results}).
    \item We propose a structured and compact template library with  around 500 thought templates curated from challenging mathematical problems. This library facilitates efficient retrieval and adaptation of relevant high-level thought templates for a series detailed reasoning steps.
    \item We develop hierarchical reinforcement learning on a sequence of high-level thought templates, to enable LLMs to generate an optimal \textit{thought template trajectory} for a series of simpler sub-problems, effectively simplifying the search space of reasoning paths.
    \item We design an new inference scaling system (in \cref{pic-inference}) by adaptively scaling thought templates for hierarchical reasoning. This system allows \method to dynamically retrieve a series of high-level templates and adaptively perform instantiated reasoning at inference time, achieving a better exploration-exploitation trade-off for robust and efficient problem-solving. Moreover, our \method contains more explainable reasoning structures than DeepSeek-R1 \citep{guo2025deepseekr1} and o3-mini.
\end{enumerate}


\section{Related Work and Discussions}
\paragraph{Learning from Preferences for Language Models}

Preference learning is critical for aligning Large Language Models (LLMs) with human expectations and perceptions. Initial approaches, building on pre-training and supervised fine-tuning (SFT), employed PPO in Reinforcement Learning from Human/AI Feedback (RLHF/RLAIF) frameworks \citep{PPO,christiano2017deep,ouyang2022training,MCTSDPO}. These approaches typically involve training a reward model on preference pairs and subsequently optimizing the LLM to maximize the learned reward. However, PPO's instability and inefficiency motivated alternative approaches like DPO \citep{DPO}, which directly optimizes a policy from paired preference data.
Subsequent research has addressed various challenges. ORPO \citep{ORPO} integrates alignment into SFT, KTO \citep{KTO} leverages pointwise data, simplifying data acquisition process. Other efforts focus on finer-grained optimization, such as Step-DPO \citep{StepDPO} and Cross-DPO \citep{yang2024supercorrect} that targets intermediate reasoning or reflection steps. SPO \citep{MinMaxPPO} employs game-theoretic concepts to address non-transitive preferences, while Multi-turn DPO \citep{DMPO} extends optimization to conversations. 
However, existing methods often rely on instance or step-level reward units, potentially failing to capture and reward the higher-level cognitive processes inherent in human problem-solving process. 
To this end, we introduce hierarchical RL-based optimization, a novel preference learning approach that encourages the model to configure a series of high-level thought templates that can handle diverse sub-tasks for complex problems, thereby promoting more human-like problem-solving strategies in LLMs.

\paragraph{Retrieval-Augmented Generation for Language Models}
Retrieval-augmented Language Models (RALMs) have become a powerful approach to mitigating hallucinations and enhancing the factual accuracy of LLMs \citep{asai2023retrieval,mialon2023augmented,shi2023replug,gao2023retrieval,zhao2024retrieval}. By retrieving relevant documents from a large-scale external knowledge source \citep{borgeaud2022improving} to inform response generation, RALMs have demonstrated superior performance in question-answering, often with fewer parameters than traditional LLMs \citep{mialon2023augmented}. Their versatility is further evidenced by successful applications across diverse tasks, including multi-modal generation and biomedical applications \citep{yasunaga2023retrieval,izacard2023atlas,wang2022retrieval,zhao2024retrieval,borgeaud2022improving,yang2023prompt}.  However, RALMs face challenges in complex reasoning tasks, such as math and code, where retrieving relevant guidelines or templates via standard embedding similarity search proves difficult. While methods like RAFT \citep{RAFT} have attempted to address this by improving retrieval relevance, respectively, their effectiveness decrease as the document size grows.  To overcome these limitations, we design a structured and compact template library for efficient and accurate retrieval, specifically targeting complex reasoning problems.

    

\paragraph{Inference Scaling for LLM Reasoning}

The auto-regressive nature of LLMs suggests that solving more complex problems inherently requires generating more tokens. Early work, such as CoT \citep{CoT}, used prompting techniques like "Let's think step by step" to break down complex reasoning tasks into simpler sub-problems, thus enhancing reasoning performance. Building on this, ToT \citep{ToT} and GoT \citep{GoT} employed different data structures to expand the reasoning space, allowing LLMs to explore multiple solution paths. Recent research \citep{inferenceScaling,snell2024scaling} has formalized the concept of inference scaling laws, which examine the trade-offs between the generation of additional tokens, and the use of various inference strategies. For instance, majority voting and best-of-N methods \citep{wang2022self,li2023making} generate multiple candidate solutions and select the best based on frequency among all the results or the reward model's evaluation.  Similarly, approaches using Monte Carlo Tree Search (MCTS) \citep{zhang2023planning,liu2024don,choi2023kcts,zhou2023language} leverage greater search and computation to improve accuracy. To enhance search accuracy, Process Reward Models (PRMs) have been introduced to select high-quality reasoning paths, with studies \citep{setlur2024rewarding,snell2024scaling,lightman2023let,luo2024improve,wang2024math} demonstrating their effectiveness, particularly in complex reasoning tasks. 
More recently, methods like BoT \citep{BoT} utilize thought templates from past reasoning processes to guide exploration, significantly improving efficiency. However, a deeper understanding of the exploration-exploitation trade-off \citep{tang2024code,setlur2024rewarding} for these template-based approaches remains an open challenge. Our work addresses this challenge by scaling an hierarchical template-augmented reasoning paradigm that significantly enhances reasoning accuracy, especially for complex tasks, while strategically balancing exploration and exploitation.

\section{\method: Scaling Thought Templates for Hierarchical LLM Reasoning}

\subsection{Constructing Structured Thought Template Library}
\label{sec-template}

Inspired by how humans utilize external resources when tackling complex reasoning problems, RAG methods enhance LLMs by enabling them to retrieve information from external sources \citep{zhao2024retrieval}. Recent Buffer of Thought (BoT) \citep{BoT} attempts to create a buffer of high-level thoughts for llm reasoning, and builds an efficient RAG reasoning system. Despite a comprehensive template library to solve similar problems, BoT still faces scalability challenges as template size grows, same as the traditional RAG systems that rely on embedding similarity to search unstructured text corpora. 

To address this, our approach focuses on constructing a \textbf{structured thought template library} that enables more precise, targeted retrieval and mitigates scalability challenges. 
To build this library, we carefully selected a wide and diverse range of challenging mathematical reasoning problems from different sources, ensuring robustness and broad applicability of our template library. 
We used an LLM to analyze the thought behind the solution and generating concise summaries of problem-solving strategies and identifying common patterns. This process yielded a collection of high-quality, solution-oriented thought templates.
Each template $T_i$ in the library is structured for efficient retrieval and application, 
where $T_{\text{nam}}$ is the \textbf{name} (e.g., "$\sqrt{R^2 - x^2}$ Type Trigonometric Substitution"), $T_{\text{tag}}$ is a set of \textbf{tags} for keyword-based retrieval (e.g., \{``Trigonometric Substitution", ``Irrational Function Optimization"\}), $T_{\text{des}}$ is a \textbf{description} of the underlying principle and applicable scenarios, $T_{\text{sco}}$ defines the \textbf{scope}, specifying the problem types it addresses,  $T_a$ is a sequence of detailed \textbf{application steps} $\{a_1, a_2, ..., a_k\}$, and $T_{\text{exa}}$ is a set of \textbf{examples} demonstrating its application. The entire library $\mathcal{D}_{\text{temp}}$ is a set of thought templates as mentioned:

\begin{equation}
\label{eq-library}
\mathcal{D}_{\text{temp}} = \{T_1, T_2, ..., T_m\}
\end{equation}

where $m$ is the total number of templates. Here we present an illustration of a thought template within our library. For the sake of brevity, some fields in the following example have been simplified. Please refer to \cref{sec-app-template-example} for more detailed examples.

\begin{tcolorbox}[title=Example Template, colback=white!5!white,colframe=black!75!black,fonttitle=\bfseries, left=1mm, right=1mm, top=1mm, bottom=1mm]
\textbf{name}: $\sqrt{R^2 - x^2}$ Type Trigonometric Substitution

\textbf{tag}: Substitution Method, Trigonometric Substitution, Irrational Function

\textbf{description}: When a radical of the form $\sqrt{R^2 - x^2}$ appears in a problem, and $|x| \le R$, consider using trigonometric substitution $x = R\sin\theta$ or $x = R\cos\theta$ to eliminate the radical, converting the irrational expression into a trigonometric expression. This allows simplification and problem-solving using the properties and identities of trigonometric functions.

\textbf{scope}: Problems involving function optimization or range, especially those involving irrational functions of the form $\sqrt{R^2 - x^2}$. Equations or inequalities containing radicals of the form $\sqrt{R^2 - x^2}$. Geometric problems related to circles.

\textbf{application steps}:

1. \textbf{Determine the range}: Based on the problem conditions, determine the range of $x$, usually $|x| \le R$.

... (Steps 2-5 omitted for brevity)

\textbf{example}:

... (Examples omitted for brevity)

\end{tcolorbox}

Efficient retrieval is facilitated by leveraging the metadata associated with each template, specifically the \textbf{name} ($n$) and \textbf{tags} ($t$), enabling quick and accurate searching based on keywords or specific problem characteristics. This structured organization, combined with rich metadata, ensures that the most relevant templates are readily available for any given problems.

\subsection{Hierarchical Reinforcement Learning on Thought Template Trajectory}
\label{sec-rewarding}
While our structured template library provides a valuable resource for reasoning, an effective method is needed to utilize this library and select the appropriate templates for handing a given problem.  To this end, we perform hierarchical reinforcement learning to train and finally obtain \textbf{\method} that can effectively plan out an optimal \textit{thought template trajectory} for a problem. We retrieve and configure a sequence of relevant templates from the library, assisting in instantiating the retrieved templates on specific sub-problems. \method acts as an experienced navigator, providing the optimal trajectory denoted as $\mathbb{T}_{\text{traj}}$ that enabling the LLM to instantiate abstract thought templates into concrete sequential problem-solving steps.

\paragraph{Structure-based Finetuning} Our hierarchical RL process begins by leveraging the structured template library $\mathcal{D}_{\text{temp}}$ to construct a knowledge-intensive training dataset $\mathcal{D}_{\text{train}}$. This dataset comprises diverse examples of template names $T_{\text{nam}}$, their associated tags $T_{\text{tag}}$, detailed descriptions of their underlying principles $T_{\text{des}}$, and a clear delineation of their applicable scopes $T_{\text{sco}}$, represented as tuples $(T_{\text{nam}}, T_{\text{tag}}, T_{\text{des}}, T_{\text{sco}})$ extracted from $\mathcal{D}_{\text{temp}}$. We then fine-tune a base LLM, denoted as $\pi$, on this dataset $\mathcal{D}_{\text{train}}$. This process equips the model with a foundational understanding of the structure, content, and intended use of each template within the library. The fine-tuning process is driven by the following optimization objective:

\begin{equation}
\label{eq-sft}
\mathcal{L}_{\text{struct}} = -\mathbb{E}_{\mathcal{D}_{\text{train}}} \left[\log \pi(T_{\text{des}}, T_{\text{sco}} | T_{\text{nam}}, T_{\text{tag}}) \right],
\end{equation}

where the objective is to maximize the likelihood of the model generating the correct description $T_{\text{des}}$ and scope $T_{\text{sco}}$ given the template name $T_{\text{nam}}$ and tags $T_{\text{tag}}$. This ensures that the fine-tuned model can effectively associate the identifying information ($T_{\text{nam}}$ and $T_{\text{tag}}$) of a template with its functional aspects ($T_{\text{des}}$ and $T_{\text{sco}}$). After fine-tuning, we denote the resulting model as $\pi_{\text{struct}}$.

\paragraph{Preference Learning on Thought Template Trajectory} Based on the finetuned LLM $\pi_{\text{struct}}$, we can further enhance its ability to plan out a sequence of high-level thought templates (\textit{i.e.}, \textbf{thought template trajectory} $\mathbb{T}_{\text{traj}}$) for an input problem $x$, associating each step with the most relevant template from the library. 
This is achieved through our preference learning on thought template trajectory. Specifically, as shown in \cref{pic-training}, given an input problem $x$, $\pi_{struct}$ first analyzes and abstracts the problem's conditional information, identifying the core mathematical concepts and relationships involved. Based on this abstract representation, the navigator $\pi_{\text{struct}}$ then configures a trajectory $\mathbb{T}_{\text{traj}} = \{s_1, s_2, ..., s_n\}$, where each $s_i$ represents a high-level step in the reasoning process, associated with a specific template name retrieved from the library which could be used to solve the problem, denoted as $T_i$. Each retrieved template $T_i$ is then instantiated with specific details from the input problem $x$ and provides fine-grained guidance to a separate inference LLM denoted as $\pi_{\text{inf}}$ to solve the problem.

To measure the effectiveness and generalization ability of a given trajectory, we utilize a set of problems $\mathcal{X}_{sim}$ that are similar to the original input problem $x$, including $x$ itself. We then use the instantiated templates along the trajectory $\mathbb{T}_{\text{traj}}$ to guide $\pi_{inf}$ in solving each problem $x_i \in \mathcal{X}_{sim}$. The average accuracy achieved by $\pi_{inf}$ across these problems serves as the \textbf{trajectory reward} $R(\mathbb{T}_{\text{traj}})$. Formally:

\begin{equation}
\label{eq-reward}
R(\mathbb{T}_{\text{traj}}) = \frac{1}{|\mathcal{X}_{sim}|} \sum_{x_i \in \mathcal{X}_{sim}} \text{Acc}(\pi_{inf}(x_i, \mathbb{T}_{\text{traj}}))
\end{equation}

where $\text{Acc}(\pi_{inf}(x_i, \mathbb{T}_{\text{traj}}))$ represents the accuracy of $\pi_{inf}$ in solving problem $x_i$ when guided by the trajectory $\mathbb{T}_{\text{traj}}$.

This reward signal is then used to construct optimization pairs, enabling us to further refine the navigator $\pi_{struct}$. To be more specific, for each input problem $x$, we sample multiple different $\mathbb{T}_{\text{traj}}$ and evaluate its quality utilizing the template trajectory reward. We define the loss function for optimizing $\pi_{struct}$ as follows:

\begin{align}
\label{eq-rewarding-loss}
\begin{split}
\mathcal{L}_{\text{TTR}}(\theta) = -\mathbb{E}_{(x, (\mathbb{T}_{\text{traj}}^+, \mathbb{T}_{\text{traj}}^-)) \sim \mathcal{D}_{pair}} \Bigg[ & \log \sigma \bigg(\beta \log \frac{\pi_{\theta}(\mathbb{T}_{\text{traj}}^+|x)}{\pi_{sft}(\mathbb{T}_{\text{traj}}^+|x)}
- \beta \log \frac{\pi_{\theta}(\mathbb{T}_{\text{traj}}^-|x)}{\pi_{sft}(\mathbb{T}_{\text{traj}}^-|x)}\bigg) \Bigg]
\end{split}
\end{align}

where $\mathcal{D}_{pair}$ is a dataset of optimization pairs. Each pair consists of an input problem $x$ and two trajectories, $\mathbb{T}_{\text{traj}}^+$ and $\mathbb{T}_{\text{traj}}^-$, where $R(\mathbb{T}_{\text{traj}}^+) > R(\mathbb{T}_{\text{traj}}^-)$.  $\pi_{\theta}$ represents the the LLM being optimized with parameters $\theta$, initialized from $\pi_{struct}$.   

\subsection{Inference Scaling with Scaling Thought Templates}
\label{sec-inference}

After hierarchical RL process, we refer to optimized navigator $\pi_{\theta}$ as \method. Then, we further design a novel inference scaling system by leveraging automatically planned trajectories and dynamically retrieved thought templates. This system, illustrated in \cref{pic-inference}, involves a multi-round interplay between the \method, a structured template library $\mathcal{D}_{\text{temp}}$, and a downstream inference LLM $\pi_{inf}$.

Given an input problem $x$,  the first task for \method is to analyze and extract the core mathematical concepts and relationships embedded within $x$. Based on this abstract representation, denoted as $a(x)$. \method then configures an optimal template trajectory $\mathbb{T}_{\text{traj}}^*$. This trajectory, represented as a sequence of steps $\mathbb{T}_{\text{traj}}^* = \{s_1^*, s_2^*, ..., s_n^*\}$, is not a rigid, pre-defined path but rather a dynamically generated plan tailored to the specific nuances of the input problem $x$. Each step $s_i^*$ within the trajectory is associated with a specific template name $T_\text{{nam}}$ and $T_\text{{tag}}$ for efficient retrieval.
\method then searches and retrieves a set of most relevant thought templates from the curated thought template library $\mathcal{D}_{\text{temp}}$.  Formally, the retrieval process can be represented as:
\begin{equation}
\label{eq:retrieval}
T_{\text{rag}} = \textit{ReasonFlux}(\{T_\text{{nam}}^i,T_\text{{tag}}^i\}_{i=1}^n, \mathcal{D}_{\text{temp}}),
\end{equation}
where $T_{\text{rag}} = \{T_1, T_2, ..., T_n\}$ is the set of $n$ retrieved templates that equals to the number of steps in the configured trajectory, and each is a structured template.

Subsequently, based on the $\mathbb{T}_{\text{traj}}^*$ and retrieved templates $T_{\text{rag}}$, \method will instruct $\pi_{inf}$ to instantiate each steps $s_i^*$ along with corresponding template $T_i$ and problem-specific details from $x$, transforming into concrete instantiated reasoning steps $\hat{s_i}$:

\begin{equation}
\label{eq-ins}
    \hat{s_i} = \pi_{inf}(x_i,s_i,T_i),
\end{equation}
where each $\hat{s_i}$ is generated based on the corresponding $s_i^*$, $T_i$, and $x$.

The interaction between \method and $\pi_{inf}$ is not a one-way process but rather in an iterative manner. After obtaining the instantiated step $\hat{s_i}$, it is then evaluated and analyzed by \method, and we represented this adjustment as process $ \delta_i =\textit{ReasonFlux}(\mathbb{T}_{\text{traj}}^*,\hat{s_i})$. Based on this evaluated result and analysis, \method decide whether to refine the trajectory, potentially adjusting subsequent steps or even retrieving alternative templates. This iterative refinement can be expressed as:

\begin{equation}
\label{eq-adjust}
\mathbb{T}_{\text{traj}}^* \leftarrow \textit{ReasonFlux}(\mathbb{T}_{\text{traj}}^*, \delta_i).
\end{equation}

This iterative feedback mechanism between \method and $\pi_{inf}$ underscores a crucial aspect of complex problem-solving: the dynamic interplay between planning and execution. By analyzing intermediate results generated during the reasoning process, \method gains valuable insights that can inform adjustments to the trajectory. This ability to refine the solution path precisely reflects how humans often uncover more efficient or effective solutions by examining partial results.  Furthermore, intermediate steps may reveal previously obscured constraints or opportunities within the problem, allowing for a more informed and targeted approach.  Therefore, the hierarchical nature of \method, enabled by this iterative refinement, is crucial for navigating the complexities of challenging reasoning tasks and achieving optimal solutions. In summary, \method achieves effective problem solving by dynamically configuring and adjusting the template trajectory based on the problem complexity, transcending the limitations of traditional inference methods and offering a more efficient and powerful reasoning framework.

\begin{table*}[htb!]
\centering
\caption{Pass@1 accuracy comparison on various mathematical reasoning benchmarks.}
\label{tab:results}
\resizebox{\textwidth}{!}{%
\begin{tabular}{lcccccc}
\toprule
Model
& MATH-500 & AIME 2024 & AMC 2023 & Olympiad Bench & Gaokao En 2023 \\
\midrule
\textbf{Frontier LLMs} & & & & & \\
GPT-4o & 76.6 & 9.3 & 47.5 & 43.3 & 67.5 \\
Claude3.5-Sonnet & 78.3 & 16.0 & - & - & - \\
GPT-o1-preview & 85.5 & 44.6 & 90.0 & - & 71.4 \\
GPT-o1-mini & 90.0 & 56.7 & 95.0 & 65.3 & 78.4 \\
\midrule
\textbf{Open-Sourced Reasoning LLMs} & & & & & \\
DeepSeek-Coder-V2-Instruct & 75.3 & 13.3 & 57.5 & 37.6 & 64.7 \\
Mathstral-7B-v0.1 & 57.8 & 0.0 & 37.5 & 21.5 & 46.0 \\
NuminaMath-72B-CoT & 64.0 & 3.3 & 70.0 & 32.6 & 58.4 \\
LLaMA3.1-8B-Instruct & 51.4 & 6.7 & 25.0 & 15.4 & 38.4 \\
LLaMA3.1-70B-Instruct & 65.4 & 23.3 & 50.0 & 27.7 & 54.0 \\
LLaMA3.1-405B-Instruct& 73.8&-&-&34.8&-\\

Qwen2.5-Math-72B-Instruct & 85.6 & 30.0 & 70.0 & 49.0 & 71.9 \\
rStar-Math&88.2&43.3&80.0&63.1&78.2\\
DeepSeek-V3 & 90.2 & 39.2 & 80.0 & 55.4 & -\\

\textbf{ReasonFlux-32B} & \textbf{91.2}& \textbf{56.7}&\textbf{85.0} &\textbf{63.3}&\textbf{83.6}\\
\midrule
& & & \textit{1.5B-Level Base Model}& &\\
Qwen2.5-Math-1.5B & 51.2&  0.0&  22.5&  16.7& 46.5\\
Qwen2.5-Math-1.5B-Instruct&  60.0&  10.0&  60.0&  38.1 & 65.5\\
\textbf{ReasonFlux-1.5B} & \textbf{70.4}& \textbf{20.0}&\textbf{72.5} &\textbf{49.0}&\textbf{76.6}\\
\midrule
& & & \textit{7B-Level Base Model}& &\\
Qwen2.5-Math-7B  &58.8 &3.3& 22.5& 21.8 &51.7\\
SuperCorrect-7B&70.2&10.0&37.5&39.0&64.0\\
Qwen2.5-Math-7B-Instruct& 82.6 &13.3& 62.5 &41.6 &66.8\\
\textbf{ReasonFlux-7B} & \textbf{88.6}& \textbf{36.7}&\textbf{80.0} &\textbf{54.8}&\textbf{80.5}\\
\midrule
& & & \textit{32B-Level Base Model}& &\\
Qwen2.5-32B-Instruct&79.4&16.5&64.0&45.3&72.1\\

QwQ-32B-preview & 90.6&50.0&75.0&-&65.3\\

Sky-T1-32B-preview&86.4&43.3&-&59.8&-\\

\textbf{ReasonFlux-32B} & \textbf{91.2}& \textbf{56.7}&\textbf{85.0} &\textbf{63.3}&\textbf{83.6}\\

\bottomrule
\end{tabular}}
\vspace{-0.2in}

\end{table*}

\section{Experiments}

\paragraph{Template Library Construction} As illustrated in \cref{sec-template}, we use Gemini-2.0 \citep{team2023gemini} to summarize and extracts high-level thoughts from  the training sets of various math datasets, such as MATH (7.5K samples) \citep{lightman2023let}, and self-curated CN high-school competition-level data (2K samples), and construct our structured thought template library (approximately 500 thought templates). We provide some template examples in \cref{sec-app-template-example}.

\paragraph{Training Details}
Due to limited GPU resources, we use Qwen2.5-32B-Instruct \citep{qwen2.5} as the base model and also adopt it as our inference LLM. In our training procedure, we \textbf{only use 8 NVIDIA A100 GPUs}, which is very cost-efficient. In the structure-based finetuning stage (\cref{sec-rewarding}), we train the initialized $\pi_{\text{struct}}$ with the training dataset $\mathcal{D}_{\text{train}}$  containing 15K samples extended from our template library $\mathcal{D}_{\text{temp}}$. We conduct the initialization training for 6 epochs using an AdamW optimizer along with the cosine learning rate scheduler.
In the template trajectory optimization process (\cref{sec-rewarding}), we train our \method with 10K collected pair-wise trajectories from MATH (7.5k), and self-curated CN high-school competition-level data (2K) for 6 epochs  using an AdamW optimizer along with cosine learning rate scheduler.



\paragraph{Evaluation Datasets} To evaluate the complex reasoning capabilities, we choose a broad set of challenging reasoning benchmarks, including MATH~\citep{lightman2023let}, AIME 2024~\citep{aime}, AMC 2023~\citep{amc}, OlympiadBench~\citep{he2024olympiadbench} and GaoKao (Chinese
College Entrance Exam) En 2023~\citep{liao2024mario}. These benchmarks comprehensively evaluate mathematical reasoning capabilities, and they are all competition-level and Olympic-level problems. Moreover,  AIME 2024 and AMC 2023 are highly challenging competition benchmarks, which are of limited sizes of test samples in AMC and AIME and the results are averaged over 16 runs.

\paragraph{Baselines} To demonstrate reasoning ability of {\method}, we compare it with two kinds of strong baseline models: \textbf{(i)} \textit{Frontier LLMs} contain GPT-4o, Claude, OpenAI o1-preview and o1-mini. We report their performance on our evaluation benchmarks by taking accuracy numbers from different public technical reports. 
\textbf{(ii)} \textit{Open-sourced superior reasoning models} contain DeepSeek-Coder-v2-Instruct, Mathstral~\citep{mathstral}, NuminaMath-72B~\citep{numina_math_datasets}, LLaMA3.1~\citep{llama3.1}, Qwen2.5-Math \citep{qwen2.5}, SuperCorrect-7B-Instruct \citep{yang2024supercorrect}, QwQ-32B-Preview \citep{qwq-32b-preview}, rStar-Math \citep{guan2025rstar} and Sky-T1-32B-Preview (distilled from QwQ-32B-Preview), and DeepSeek-V3 \citep{deepseekv3}, which are widely used and followed open-sourced reasoning models. Both kinds of baselines represent the highest level of mathematical reasoning currently available.

\subsection{Results on Challenging Reasoning Benchmarks}
\cref{tab:results} shows the final results of our \method with a comprehensive comparison to SOTA reasoning models. We find that our \method-32B consistently outperforms both frontier LLMs and open-sourced reasoning LLMs on most challenging mathematical benchmarks, achieving new SOTA performances with only 32B-level parameters. 
More specifically, on the MATH benchmark, \method achieves 91.2\% of accuracy, \textbf{surpassing frontier reasoning models o1-preview by 6.7\%}, and current SOTA-level open-source LLMs \textbf{with only 32B parameters}.
On the AIME 2024 benchmark, \method consistently demonstrates its extrodinary reasoning capabilities with 56.7\% accuracy, \textbf{significantly surpassing o1-preview and DeepSeek-V3 by 27\% and 45\%}, respectively, and matching the performance of the proprietary OpenAI o1-mini.
On the AMC 2023 benchmark, our method, \method, maintains its position within the top tier of all reasoning LLMs with 85.0\% accuracy, significantly outperforming other open-source LLMs while achieving performance comparable to proprietary LLMs. This further validates the effectiveness of our approach in mathematical reasoning and underscores its substantial potential for further development and application. We provide some reasoning details in \cref{sec-app-detailed-reasonflow}. 

Beyond above well-known benchmarks, \method-32B also demonstrates impressive generalization and effectiveness on other challenging datasets. Notably, it achieves a 63.3\% accuracy \textbf{on OlympiadBench surpassing DeepSeek-V3 by 14\%}, and an 83.6\% accuracy \textbf{on the Chinese College Entrance Mathematics Exam (Gaokao) surpassing o1-mini by 7\%}. These results are particularly noteworthy because our template library was constructed primarily from publicly available datasets, the same template library was used consistently across all evaluation processes. This consistent strong performance across diverse and challenging mathematical reasoning tasks, ranging from competition-level problems to standardized exams, provides compelling evidence for the robust generalization ability and effectiveness of \method. It underscores the power of our template-driven approach to capture and apply underlying mathematical principles, regardless of the specific format or context of the problem.

\paragraph{Generalizing to Different Base Models} From \cref{tab:results}, we also observe that our \method can achieve consistent and significant improvement across all evaluation benchmarks when using different base models as both navigator and inference LLM. Notably, our \method usually achieves even surpasses the reasoning accuracy of the models in next level. These phenomenons demonstrate both effectiveness and generalization ability of our \method. 
\begin{table*}[h]
\centering
\caption{Generalization ability of our thought templates with different base LLMs on a series of similar mathematical problems.}
\begin{tabular}{lccc} 
\hline 
Model & direct reasoning (\%) & with Template (\%) \\ 
\hline 
\textbf{Llama-3.1-8B-Instruct} & 47.6 & 75.1 (\textbf{+27.5}) \\
\textbf{Qwen2.5-7B-Instruct} & 59.2 & 82.7 (\textbf{+23.5}) \\
\textbf{Qwen2.5-Math-7B-Instruct} & 66.5 & 88.4 (\textbf{+21.9}) \\
\textbf{Llama-3.1-70B-Instruct} & 67.4 & 91.2 (\textbf{+23.8}) \\
\textbf{Qwen2.5-32B-Instruct} & 69.2 & 94.3 (\textbf{+25.1}) \\
\textbf{Qwen2.5-Math-32B-Instruct} & 71.1 & 95.9 (\textbf{+24.8}) \\
\hline 
\end{tabular}
\vspace{-0.2in}
\label{tab:generalization}
\end{table*}

\begin{figure*}[tp]
    \centering
    
    \includegraphics[width=\textwidth]{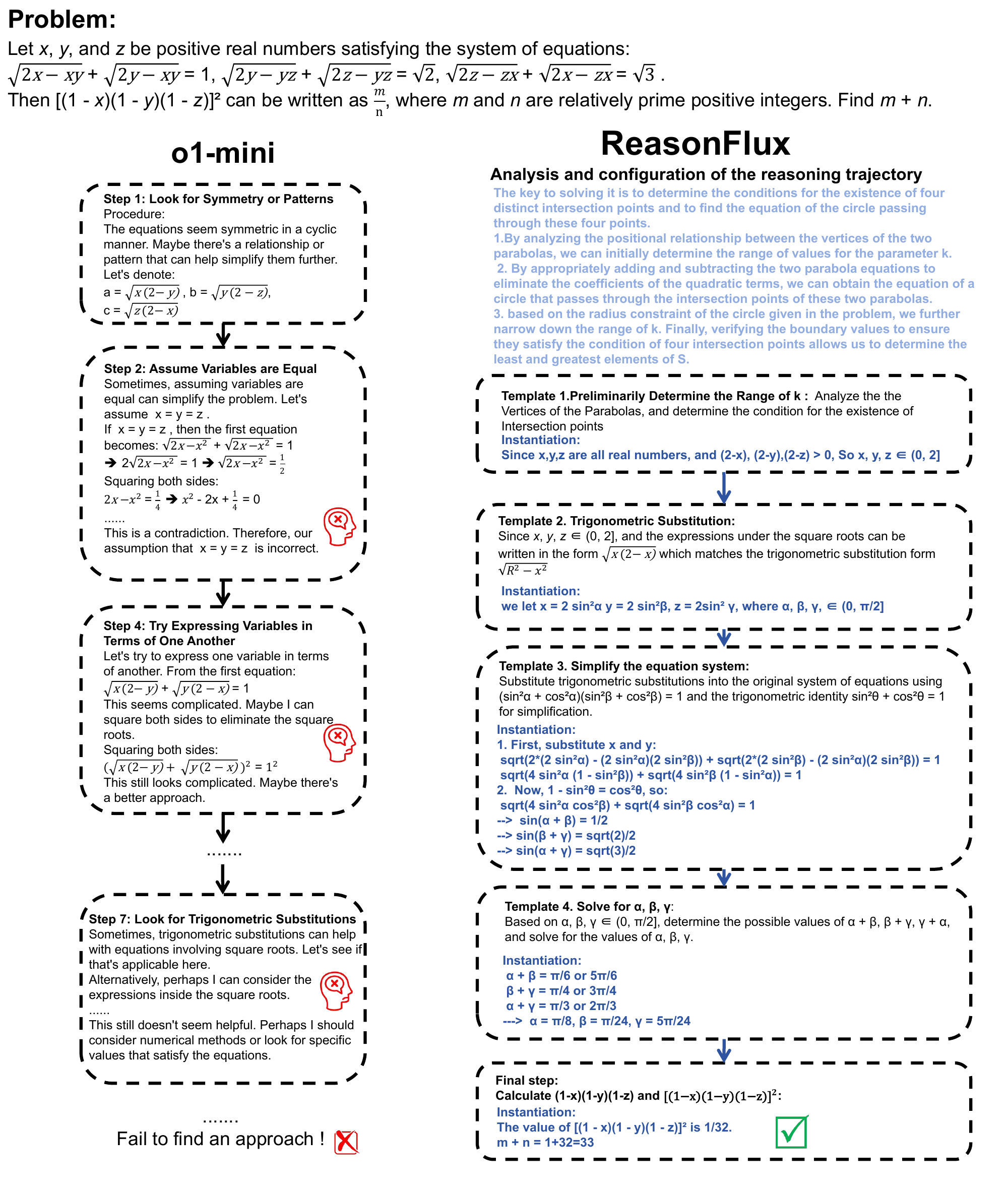}
    \vspace{0.1in}
    \caption{Comprasion between o1-mini and \method.}
    \vspace{-0.2in}
    \label{pic-app-deepseek}
\end{figure*}

    

\subsection{Generalization Ability of Structured Template Library}
\label{sec-app-template-general}
We presents additional experiments on MATH benchmark designed to evaluate the generalization ability of our structured template library. To achieve this, we randomly sampled 100 templates from the library, each paired with its corresponding example problem. Subsequently, we employed o1-preview to generate 50 variant problems for each example. These variants were carefully constructed to ensure they differed from the original examples while still assessing the same underlying knowledge and skills.

We then used these templates as in-context examples to guide different LLMs during inference on the generated variant problems. We compare the average accuracy between our template augmented reasoning and direct reasoning (i.e., solving the problems without template). As illustrated in \cref{tab:generalization}, our template-augmented approach significantly improves the reasoning accuracy of different base models compared to direct reasoning. This demonstrates the ability of our structured templates to generalize effectively across a range of similar problems, rather than being limited to specific instances. Furthermore, we observed that smaller-sized LLMs, when guided by our templates, were able to outperform larger-sized LLMs employing direct reasoning. This finding underscores the effectiveness and high quality of our structured template library.

\subsection{Reasoning Flows over Planned Template Trajectory}
\label{sec-app-detailed-reasonflow}
We showcase detailed examples of our reasoning flows, as depicted in \cref{pic-app-deepseek}, when tackling challenging mathematical problems.  Specifically, ReasonFlux begins by meticulously observing and analyzing the input problem, engaging in deep thought to explore potential solution pathways. Based on this initial assessment, ReasonFlux intelligently configures a dynamic reasoning trajectory, strategically retrieving relevant templates from our structured template library to guide each logical step.  Then, ReasonFlux initiates an interactive instruction with the inference LLM, guiding it to follow the prescribed trajectory and execute the reasoning process along the trajectory. Crucially, the results obtained from preceding steps are seamlessly integrated as contextual information, informing and conditioning the subsequent steps. Compare to conventional self-explore and reasoning paradigm, our method could consistently improve the reasoning accuracy and efficiency. Moreover, our \method contains more explainable reasoning structures than recent powerful models like DeepSeek-R1 \citep{guo2025deepseekr1} and o3-mini.

\subsection{Inference Scaling Laws for Template-Augmented Reasoning} 
\begin{figure*}[htb!]
    \centering
    \includegraphics[width=\textwidth]{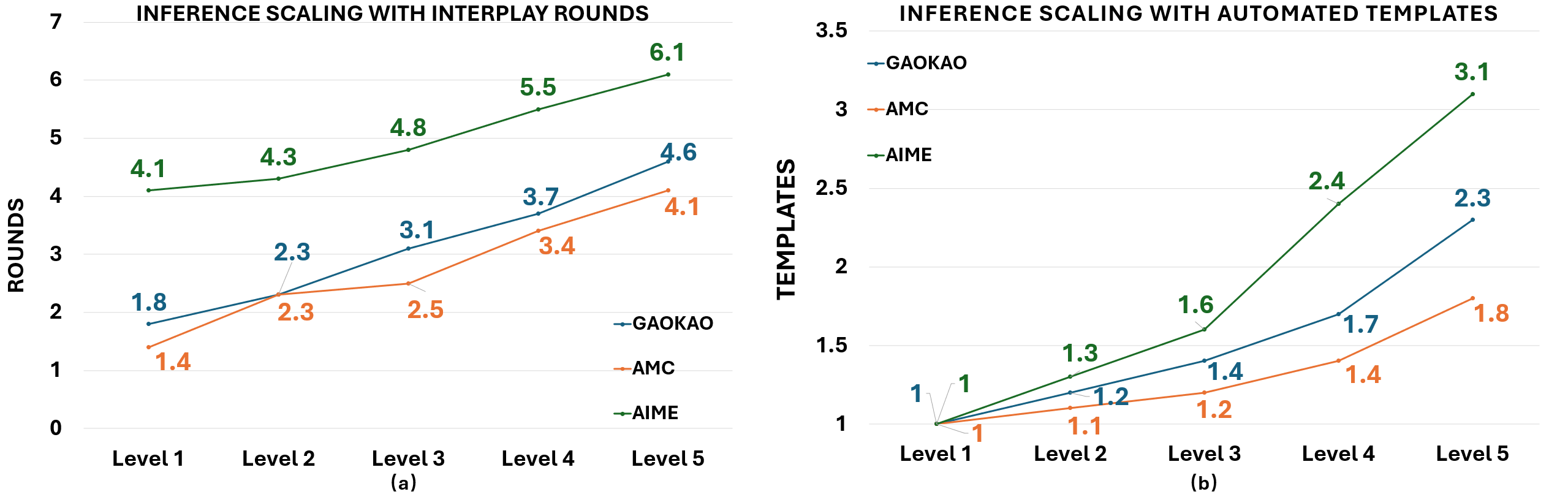}
    \vspace{-0.2in}
    \caption{\textbf{Inference scaling laws for template-augmented reasoning in \method.} (a) Scaling interplay rounds between planning and instantiation with increased  level of problem complexity.   (b) Scaling retrieved templates with increased level of problem complexity.}
    \label{pic-scaling}
\end{figure*}

Different from traditional inference scaling with Best-of-$N$ and Majority Voting \citep{inferenceScaling}, our \method owns a specific interplay-based scaling mechanism. In order to provide a comprehensive understanding of how \method automatically trade off between cost and performance. As shown in \cref{pic-scaling}, we demonstrate (i) how number of retrieved templates adaptively scales with increased problem complexity and (ii) how rounds of interplay between \method and inference LLMs adaptively scales with increased problem complexity. From the results, we can observe that our \method can effectively capture the complexity of input problems, and plan out reasonable template trajectories with appropriate interplay rounds. 
Utilizing more fine-grained thought templates may boost the scaling effect of our \method, and we leave this exploration for future work.

\begin{figure*}[htb!]
    \centering
    \includegraphics[width=0.7\textwidth]{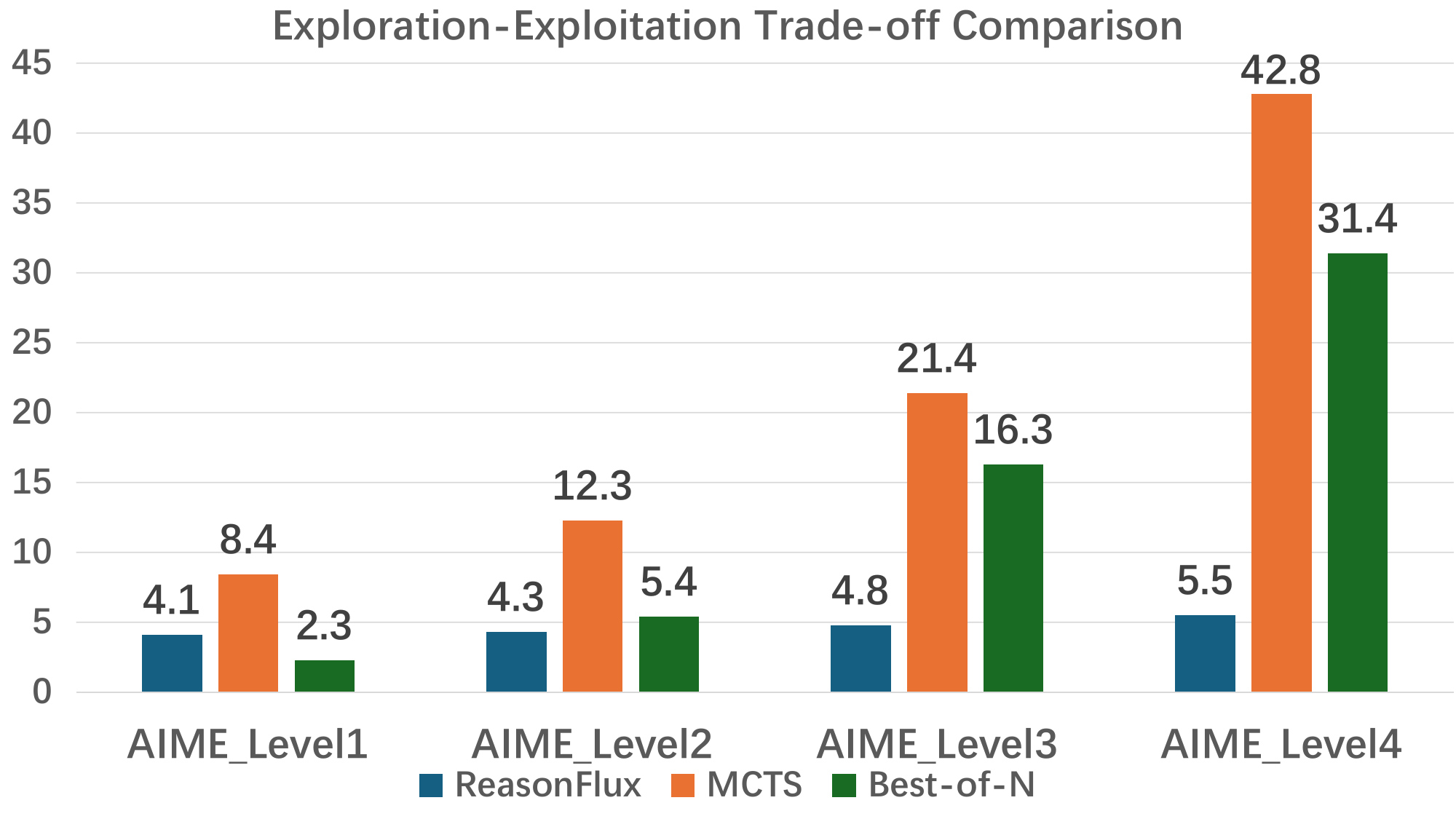}
    \vspace{-0.1in}
    \caption{\textbf{Exploration-Exploitation Trade-off Comparison between different reasoning strategies.} Here we experiment with a diverse set of 200 problems sourced from the AIME competitions spanning 1983 to 2023, divided into four difficulty levels. We test the average exploration cost of ReasonFlux (number of interplay rounds), MCTS (number of reasoning steps) and Best-of-N (number of reasoning trajectories).}
    \label{pic-trade-off}
\end{figure*}

\subsection{Better Exploration-Exploitation Trade-off}
To evaluate the exploration-exploitation trade-off of different reasoning strategies, we conducted an ablation study comparing our proposed interplay method against Best-of-N and MCTS. Each method exhibits a distinct approach to navigating the reasoning space. Best-of-N constructs multiple reasoning trajectories to identify the optimal path, while MCTS iteratively explores the most promising next step during the problem-solving process. 

Our method formulates a potential reasoning trajectory and then guides the interactive process with the inference LLM for iterative refinement and adjustments.
To ensure a fair comparison, we introduce a unified metric termed "exploration-exploitation cost." This metric quantifies the number of exploration attempts required by each method to correctly solve a given problem. For our method, this denotes the number of interactions between ReasonFlux and the inference LLM. For MCTS, it is represented by the iteration time, and for Best-of-N, it denotes the total number of sampled trajectories.

As illustrated in \cref{pic-trade-off}, both MCTS and Best-of-N exhibit an increasing exploration-exploitation cost as problem difficulty escalates. In contrast, our method maintains a consistently lower and more stable exploration cost across all difficulty levels.
This superior efficiency of our method can be attributed to the effectiveness of our structured template library. This high-quality library effectively refines the search space, facilitating the identification of correct reasoning paths. Furthermore, the high quality and generalization ability of the templates (experimental analysis in \cref{sec-app-template-general}) within the library allows for effective exploitation, guiding the Inference LLM towards accurate and efficient reasoning.  Consequently, our approach demonstrates a more balanced and efficient exploration-exploitation trade-off compared to Best-of-N and MCTS.


\section{Conclusion}
In this work, we present \method, a new hierarchical LLM reasoning framework that adaptively scales fundamental and essential thought templates for simplifying the search space of complex reasoning, and outperforming the mathematical reasoning capabilities of powerful LLMs like OpenAI o1-preview and DeepSeek V3. 
We introduces a structured and compact thought template library, hierarchical reinforcement learning on thought template trajectory and a brand new inference scaling system. Extensive experiments across different
challenging math benchmarks demonstrate the superiority of \method. We also reveal some
key findings, including the scaling laws for our template-augmented reasoning and the superior exploration-exploitation trade-off of our \method over previous reasoning strategies.



\nocite{langley00}

\bibliography{example_paper}
\bibliographystyle{icml2025}

\newpage
\appendix
\onecolumn

\section{More Examples of Structured Template Library}
\label{sec-app-template-example}
In this section, we present a more detailed and diverse collection of supplementary examples for \cref{sec-template}, showcasing our meticulously designed structured templates. These examples span a range of template types, demonstrating the versatility and applicability of our approach. The template types include: 1) \textbf{Problem-Solving Methods}, which provide step-by-step procedures for tackling specific problem types; 2) \textbf{Secondary Mathematical Conclusions}, which encapsulate derived mathematical results that can be applied to various problems; 3) \textbf{Property \& Theorem} that highlight essential mathematical properties and theorems; 4) \textbf{Knowledge Application} templates that demonstrate the application of specific mathematical concepts and techniques; and 5) \textbf{Important Formulas and Rules} templates, which offer concise summaries of crucial formulas and rules for quick reference and application.

To emphasize the structure and facilitate comprehension, each template is designed to contain two kinds of data: i) \textbf{Template Metadata} and ii) \textbf{Template Content}. The Template Metadata provides concise information about the template, including its \textbf{name $T_\text{{nam}}$, relevant knowledge tags $T_\text{{tag}}$, a brief description $T_\text{{des}}$, and typical application scenarios $T_\text{{sco}}$}. This section serves as a quick reference guide, enabling LLMs to efficiently locate and identify templates relevant to their needs. The Template Content delves into the core of the template, presenting the detailed reasoning flow and a concrete example illustrating its application. \textbf{The reasoning flow corresponding to the application steps $T_a$ and the example application corresponding to $T_\text{exa}$ in \cref{sec-template}}, which outlines the logical steps or procedures involved in utilizing the template, while the example provides a practical demonstration of how the template can be applied to solve a specific problem. This two-part structure enhances clarity and allows for both quick retrieval and in-depth understanding of each template.

The following examples have been carefully selected to provide a comprehensive overview of the capabilities of our structured template library.  Through these examples, we aim to more comprehensive overview of our structured templates, and demonstrate the effectiveness of our structured templates in promoting organized thinking, facilitating problem-solving, and ultimately enhancing mathematical understanding of LLMs.

\begin{tcolorbox}[
    title=(\uppercase\expandafter{\romannumeral 1}) Template ( Problem-Solving Method) : Five-Step Method for Solving Absolute Value Inequalities,
]

\textbf{Template Name:} Five-Step Method for Solving Absolute Value Inequalities

\textbf{Knowledge Tag:} Absolute Value Inequalities, Solving Inequalities, Combining Numerical and Graphical Methods

\textbf{Description:} This template provides a structured approach to solving absolute value inequalities using various strategies, with a focus on the squaring method and the zero-point interval method.

\textbf{Application Scenario:} Applicable to absolute value inequalities of the form $|x-a|>b$, $|ax+b|<c$, $|f(x)|>|g(x)|$, etc. Particularly suitable for complex cases involving multiple absolute value symbols or requiring interval discussions.

\textbf{Reasoning Flow:}
\begin{enumerate}
    \item Standardize the inequality to ensure the right side is non-negative (e.g., $|x-1| > |2x+3|$).
    \item Choose a solution strategy (Step 3 will present two options).
    \item \textbf{Solve using one of the following methods:}
    \begin{enumerate}
        \item[(a)] \textbf{Squaring Method:}
        \begin{enumerate}
            \item[(i)] Rearrange to the form $A^2 > B^2$.
            \item[(ii)] Expand and simplify into a polynomial inequality.
            \item[(iii)] Factor, find the roots, and use a number line to determine the solution set.
        \end{enumerate}
        \item[(b)] \textbf{Interval Method:}
        \begin{enumerate}
            \item[(i)] Mark the zero points of each absolute value expression (e.g., $x=1$ and $x=-1.5$).
            \item[(ii)] Divide the number line into intervals (e.g., $x \le -1.5$, $-1.5 < x < 1$, $x \ge 1$).
            \item[(iii)] Rewrite the inequality without absolute value signs within each interval.
            \item[(iv)] Solve the inequality in each interval and find the intersection with the interval.
        \end{enumerate}
    \end{enumerate}
    \item Verify whether the endpoint values satisfy the original inequality.
    \item Combine the solution sets from each interval, expressing the final result using set notation. (If using the interval method)
\end{enumerate}
\textbf{Example Application:}

\textbf{Problem:} Solve the inequality $|x-1| > |2x+3|$.

\textbf{Solution Process:}

\underline{Using Squaring Method (Step 3a):}
\begin{enumerate}
    \item Square both sides: $(x-1)^2 > (2x+3)^2$
    \item Expand and simplify: $x^2-2x+1 > 4x^2+12x+9 \rightarrow -3x^2-14x-8 > 0$
    \item Factor: $-(3x+2)(x+4) > 0 \rightarrow (3x+2)(x+4) < 0$
    \item Find the roots and use a number line: $x=-4$, $x=-\frac{2}{3}$ → Solution set: $(-4,-\frac{2}{3})$
\end{enumerate}

\underline{Using Interval Method (Step 3b):}

To better present our templates, we have omitted some examples that were too long. ......
\end{tcolorbox}

\begin{tcolorbox}[
    title= (\uppercase\expandafter{\romannumeral 2}) Template (Secondary Conclusion) : Application of the Inequality of Arithmetic and Geometric Means for Three and n Variables,
]

\textbf{Template Name:} Application of the Inequality of Arithmetic and Geometric Means for Three and n Variables

\textbf{Knowledge Tag:} Inequality of Arithmetic and Geometric Means, Three-Variable Inequality, n-Variable Inequality, Inequality Proof

\textbf{Description:} Extends the two-variable inequality of arithmetic and geometric means to three and n variables, suitable for handling the relationship between the sum and product of multiple positive numbers. The core formulas are: for three variables, $a^3+b^3+c^3 \ge 3abc$; for $n$ variables, the arithmetic mean is greater than or equal to the geometric mean.

\textbf{Application Scenario:} Used when there are three or more positive variables in the problem, and it is necessary to compare the relationships between sum, product, sum of squares, etc. Especially suitable for proving inequalities with multiple variables or finding the maximum/minimum values.

\textbf{Reasoning Flow:}
\begin{enumerate}
    \item Confirm that all variables are positive (ensure this through the problem's conditions or transformations if necessary).
    \item If it is a three-variable case, directly apply $a^3+b^3+c^3 \ge 3abc$ (equality holds if and only if $a=b=c$).
    \item If it is an $n$-variable case, apply the inequality of arithmetic and geometric means:
    \[\frac{a_1+a_2+...+a_n}{n} \ge \sqrt[n]{a_1a_2...a_n}\]
    (equality holds if and only if $a_1=a_2=...=a_n$).
    \item Transform the original expression into the standard form above through algebraic manipulations (such as grouping, factoring, completing the square, etc.).
    \item Combine with known conditions (such as $abc=1$) to substitute and simplify to find the maximum/minimum value.
    \item Verify that the condition for equality holds satisfies the problem's constraints.
\end{enumerate}

\textbf{Example Application:}

\textbf{Problem:} Given that $a$, $b$, and $c$ are positive numbers and $abc=1$, prove that $(a+b)^3+(b+c)^3+(c+a)^3 \ge 24$.

\textbf{Solution:}
\begin{enumerate}
    \item Confirm $a, b, c > 0$ and $abc=1$.
    \item Apply the three-variable inequality to each term in parentheses: $(a+b)^3 \ge 8ab(a+b)/8$ (needs to be adjusted to fit the form).
    \item Better solution: Directly apply $a^3+b^3+c^3 \ge 3abc$.
    \[ \because (a+b)^3+(b+c)^3+(c+a)^3 \ge 3(a+b)(b+c)(c+a) \]
    \item Apply the two-variable inequality of arithmetic and geometric means to $(a+b)(b+c)(c+a)$:
    \[ (a+b) \ge 2\sqrt{ab}, (b+c) \ge 2\sqrt{bc}, (c+a) \ge 2\sqrt{ca} \]
    \[ \therefore \text{The product} \ge 8\sqrt{a^2b^2c^2}=8abc=8 \]
    \item Substitute to get the original expression $\ge 3 \times 8 = 24$.
    \item Verify the equality condition: Equality holds if and only if $a=b=c=1$.
\end{enumerate}
\end{tcolorbox}

\begin{tcolorbox}[
    title=(\uppercase\expandafter{\romannumeral3}) Template (Property Theorem) : Extremum Value Theorem,
]

\textbf{Template Name:} Extremum Value Theorem

\textbf{Knowledge Tag:} Inequality of Arithmetic and Geometric Means, Extremum Value Theorem, Product is Maximum when Sum is Constant, Sum is Minimum when Product is Constant

\textbf{Description:} When the product or sum of two positive numbers $x$ and $y$ is a constant, their sum or product has an extremum value: when the product is constant, the sum has a minimum value; when the sum is constant, the product has a maximum value. Equality holds if and only if $x = y$.

\textbf{Application Scenario:} Suitable for finding the maximum/minimum value of the sum or product of two positive variables, especially when the product or sum of one of the expressions is a constant. For example: rectangle perimeter/area problems, function optimization problems, etc.

\textbf{Reasoning Flow:}
\begin{enumerate}
    \item Confirm that variables $x$ and $y$ are both positive.
    \item Determine if there is a constant product $xy = P$ or a constant sum $x+y = S$ in the problem.
    \item If the product is a constant $P$, then the minimum value of the sum $x+y$ is $2\sqrt{P}$ (when and only when $x = y$).
    \item If the sum is a constant $S$, then the maximum value of the product $xy$ is $\frac{S^2}{4}$ (when and only when $x = y$).
    \item Verify that the condition for equality holds satisfies the problem's requirements (e.g., the actual range of values for $x$ and $y$).
\end{enumerate}

\textbf{Example Application:}

\textbf{Problem:} What is the minimum value of the function $y = \frac{x^4 - 5x^2 + 1}{x^2 - 5}$ ($x^2 > 5$)?

\textbf{Solution:}
\begin{enumerate}
    \item Confirm the variable is positive: $x^2 > 5 \Rightarrow x^2 - 5 > 0$.
    \item Transform the function: $y = x^2 + \frac{1}{x^2 - 5} -5$.
    \item Let $a = x^2 - 5 > 0$, then $y = a + \frac{1}{a}$.
    \item Apply the Extremum Value Theorem: $a + \frac{1}{a} \ge 2\sqrt{a \cdot \frac{1}{a}} = 2$ (when and only when $a = \frac{1}{a} \Rightarrow a = 1$).
    \item Therefore $y \ge 2 + 5 = 7$, when and only when $x^2 - 5 = 1 \Rightarrow x = \pm\sqrt{6}$, the equality holds.
\end{enumerate}
\textbf{Answer:}  7
\end{tcolorbox}

\clearpage





\begin{tcolorbox}[
    title=(\uppercase\expandafter{\romannumeral4}) Template (Knowledge Application) : Analyzing the Parity and Symmetry of Trigonometric Functions Using Reduction Formulas,
]

\textbf{Template Name:} Analyzing the Parity and Symmetry of Trigonometric Functions Using Reduction Formulas

\textbf{Knowledge Tag:} Reduction Formulas, Parity, Symmetry, Properties of Trigonometric Functions

\textbf{Description:} This template guides the analysis of the parity and symmetry of complex trigonometric functions by transforming them into standard forms using reduction formulas, aiding students in systematically solving related problems.

\textbf{Application Scenario:} Applicable to determining the parity of trigonometric functions, identifying the symmetry centers or axes of function graphs, and solving for parameters (e.g., phase angle $\phi$). This method is useful when encountering functions of the form $y=A\sin(\omega x+\phi)$ or $y=A\cos(\omega x+\phi)$.

\textbf{Reasoning Flow:}
\begin{enumerate}
    \item Transform the target trigonometric function into a standard sine or cosine form using reduction formulas. For example, use $\sin(x+\frac{\pi}{2})=\cos x$ to convert a cosine function to a sine form.
    \item Determine the function's parity based on the definition of odd and even functions. An odd function satisfies $f(-x)=-f(x)$, and an even function satisfies $f(-x)=f(x)$.
    \item If symmetry is involved, determine the expressions for the symmetry axes or centers. For example, the symmetry axes of the sine function are $x=\frac{\pi}{2}+k\pi$, and the symmetry centers are $(k\pi,0)$.
    \item Compare the transformed function with the standard form and solve the equation to find the unknown parameters (e.g., $\phi$). For example, set the phase angle to satisfy the condition for an odd function, $\phi=k\pi$.
    \item Verify the solution's validity, ensuring it conforms to the original function's domain and fundamental properties.
\end{enumerate}

\textbf{Example Application:}

\textbf{Problem:} Given that the function $y=\sqrt{2} \sin(x+\phi)$ is an odd function, find the possible values of $\phi$.

\textbf{Solution Steps:}
\begin{enumerate}
    \item Based on the definition of an odd function, we have $\sqrt{2} \sin(-x+\phi) = -\sqrt{2} \sin(x+\phi)$.
    \item Expand the left side: $\sin(-x+\phi) = \sin\phi\cos x - \cos\phi\sin x$.
    \item Simplify the right side: $-\sin(x+\phi) = -\sin x\cos\phi - \cos x\sin\phi$.
    \item Compare the coefficients on both sides of the equation: $\sin\phi = -\sin\phi$ and $-\cos\phi = -\cos\phi$.
    \item Solve for $\phi$: $\sin\phi=0 \Rightarrow \phi=k\pi$ ($k\in\mathbb{Z}$).
\end{enumerate}
\end{tcolorbox}
\clearpage

\begin{tcolorbox}[
    title=(\uppercase\expandafter{\romannumeral5}) Template (Important Formulas/Rules) : Distance Formulas and Their Applications,
]

\textbf{Template Name:} Distance Formulas and Their Applications

\textbf{Knowledge Tag:} Distance Between Two Points, Distance from a Point to a Line, Distance Between Parallel Lines

\textbf{Description:} This template includes formulas for calculating three types of distances: the distance between two points, the distance from a point to a line, and the distance between two parallel lines. These formulas are core tools for solving distance problems in analytic geometry.

\textbf{Application Scenario:} This template can be applied when it is necessary to calculate the geometric distance between two points, the perpendicular distance from a point to a line, or the fixed distance between two parallel lines. It is commonly used in scenarios such as calculating the area of geometric figures, analyzing positional relationships, and solving symmetry problems.

\textbf{Reasoning Flow:}
\begin{enumerate}
    \item Step 1: Identify the type of problem (distance between two points / distance from a point to a line / distance between parallel lines).
    \item Step 2: Distance between two points formula: $|P_1P_2|=\sqrt{(x_2-x_1)^2+(y_2-y_1)^2}$, substitute the coordinates directly to calculate.
    \item Step 3: Distance from a point to a line formula: $d=\frac{|Ax_0+By_0+C|}{\sqrt{A^2+B^2}}$, ensure the line equation is in the general form $Ax+By+C=0$.
    \item Step 4: Distance between parallel lines formula: $d=\frac{|C_1-C_2|}{\sqrt{A^2+B^2}}$, both line equations must be in the form $Ax+By+C_1=0$ and $Ax+By+C_2=0$ with the same coefficients $A$ and $B$.
    \item Step 5: Handle special cases (e.g., projection distance on coordinate axes, distance transformation in symmetry problems).
\end{enumerate}

\textbf{Example Application:}

\textbf{Problem:} Given that the line $l_1: mx+2y-4-m=0$ has equal intercepts on the x-axis and y-axis, find the distance between $l_1$ and $l_2: 3x+3y-1=0$.

\textbf{Solution Steps:}
\begin{enumerate}
    \item From equal intercepts, we get $\frac{m+4}{m}=\frac{m+4}{2} \Rightarrow m=2$.
    \item Convert $l_1$ to the general form $2x+2y-6=0 \Rightarrow x+y-3=0$.
    \item Align coefficients: Rewrite $l_1$ as $3x+3y-9=0$ to match the coefficients of $l_2$.
    \item Apply the parallel lines distance formula $d=\frac{|-1-(-9)|}{\sqrt{3^2+3^2}}=\frac{8}{3\sqrt{2}}=\frac{4\sqrt{2}}{3}$.
\end{enumerate}
\textbf{Answer:} $\frac{4\sqrt{2}}{3}$
\end{tcolorbox}

\clearpage


\end{document}